\documentclass[a4paper,10pt]{article_cogis}
\usepackage{times}

\usepackage[latin1]{inputenc}
\usepackage{graphics}
\usepackage{latexsym}
\usepackage{amsmath}
\usepackage{amssymb}
\usepackage{amsfonts}
\usepackage{subfigure}
\usepackage{epsfig}

\usepackage{color}

\title{\textbf{Adaptative combination rule and proportional conflict redistribution rule for information fusion}}
\date{}

\twocolumn
\newcommand{\D}{\displaystyle} 
\newcommand{\ie}{{\it i. e. }}

\begin{document}
\pagestyle{empty}
\author{{\large{M. C. Florea$^1$, J. Dezert$^2$, P. Valin$^3$, F. Smarandache$^4$, Anne-Laure Jousselme$^3$}}\\
\normalsize $^1$ Radiocommunication \& Signal Processing Lab., Laval University, Sainte-Foy, QC, G1K 7P4, Canada.\\
\normalsize $^2$ ONERA, 29 Avenue de la Division Leclerc, 92320 Ch\^atillon, France.\\
\normalsize $^3$Defence R\&D Canada - Valcartier, 2459 Pie-XI Blvd. North Val-B\'elair, Qu\'ebec, G3J 1X5, Canada.\\
\normalsize $^4$Department of Mathematics, University of New Mexico, Gallup, NM 87301, U.S.A.
}


\maketitle

\thispagestyle{empty}

\noindent
{\bf{Abstract}}: This paper presents two new promising combination rules for the fusion of uncertain and potentially highly conflicting sources of evidences in the theory of belief functions established first in Dempster-Shafer Theory (DST) and then \hskip 0.1cm  recently \hskip 0.1cm extended \hskip 0.1cm  in \hskip 0.1cm Dezert-Smarandache Theory (DSmT). Our work is to provide here new issues to palliate the well-known limitations of Dempster's rule and to work beyond its limits of applicability. Since the famous Zadeh's criticism of Dempster's rule in 1979, many researchers have proposed new interesting alternative rules of combination to palliate the weakness of Dempster's rule in order to provide acceptable results specially in highly conflicting situations. In this work, we present two new combination rules: the class of Adaptive Combination Rules (ACR) and a new efficient Proportional Conflict Redistribution (PCR) rule. Both rules allow to deal with highly conflicting sources for static and dynamic fusion applications. We present some interesting properties for ACR and PCR rules and discuss some simulation results obtained with both rules for Zadeh's problem and for a target identification problem.\\


\section{Introduction}

Beside Zadeh's Fuzzy Set Theory (FST) \cite{Zadeh_1965}, Dempster-Shafer Theory (DST) \cite{Shafer_1976,Yager_1994} is one of most major paradigm shifts for reasoning under uncertainty. DST uses Dempster's rule to combine independent pieces of information (called sources of evidence) but this rule has been strongly criticized (and still is) in literature \cite{Zadeh_1979,Dubois_1986c,Yager_1987} because of its unexpected behaviour which can both reflect the minority opinion in some cases and provide counter-intuitive results when combining highly conflictual information as proved by Zadeh \cite{Zadeh_1979}. Some authors argue that the {\it{unexpected}} behaviour of Dempster's rule is a false problem since the reason for the counter-intuitive results comes from an improper use of this rule \cite{Voorbraak_1991,Haenni_2002,Liu_2000,Haenni_2005} and so these authors emphasize the limits of applicability of the Dempster's rule itself making DST less attractive.
%
 %
 The argument in favour of Dempster's rule is that if the initial conditions are respected and if the problem is well modelized, then Dempster's rule provides valid results. Such an argumentation is however not totally convincing since usually proponents of Dempster's rule only circumvent Zadeh's problem by changing it through more or less well justified modifications rather than solving it, and fundamentally and numerically the problem with Dempster's rule as clearly stated by Zadeh still remains open forever. Actually in many cases - specially those involving human experts - sources of evidence provide opinions or beliefs from their own limited sensing abilities, experience, knowledge with their own interpretation and understanding of the given problem and even sometimes with conflicting interests or purposes.  One has moreover not necessarily access to the quality or reliability of sources to discount them because some problems are not repeatable and we can never assess the quality of an expert facing a new problem that has never occurred in the past. There is no 100\% warranty beforehand that a complex fusion system will never fall into Zadeh's paradox \cite{DSmTBook_2004a}.
%
 %
Actually Dempster's rule appears to be satisfactory only in situations with high beliefs and low conflict, when sources agree almost totally which is rarely the case in practice. In all other cases, better alternatives to Dempster's rule have to be found to palliate its drawbacks. Since in military real-time systems, one never knows beforehand if the sources of information will be in low conflict or not, it is preferable to switch directly towards one of efficient alternative rules proposed in the literature so far \cite{Dubois_1988,Yager_1983,Yager_1987,Inagaki_1991,Lefevre_2002,Sentz_2002,Daniel_2003,Lefevre_2003,DSmTBook_2004a}. In practice the conditions of applicability of Dempster's rule (independence of homogeneous sources working on the same exhaustive and exclusive frame of discernment) are restrictive and too difficult to satisfy. Thus, the DST was extended to new more flexible theories in order to cope with an unknown and unpredictable reality. Among them, the Transferable Belief Model (TBM) of Smets and Kennes \cite{Smets_1990,Smets_1993,Smets_1994} which, by the open-world assumption, refutes the exhaustivity constraint on the frame of discernment $\Theta$ and the underlying probability model. The TBM allows to consider elements outside of $\Theta$, all represented by the empty set. More recently the Dezert-Smarandache Theory (DSmT) \cite{DSmTBook_2004a} has been developed to deal with (highly) conflicting imprecise and uncertain sources of information. DSmT provides a general framework to work with any kind of models (free or hybrid models as well as Shafer's model) and for static or dynamic fusion applications (\ie applications where the model and/or the frame are changing with time). When working with the free model, DSmT refutes the exclusivity constraint on the frame of discernment, allowing new elements than those initially considered to appear. In these two frameworks (TBM and DSmT free-based model), the conflict is no more a problem. DSmT however allows to include if necessary (depending on the application) some integrity constraints (non existential or exclusive constraints) in the modeling and propose a new hybrid rule (called DSmH) of combination for re-assignment of the conflicting mass.\\

In \hskip 0.1cm this \hskip 0.1cm paper, we \hskip 0.1cm present \hskip 0.1cm two \hskip 0.1cm new \hskip 0.1cm combination \hskip 0.1cm rules \hskip 0.1cm called ACR (Adaptive Combination Rule) \cite{Florea_2004,Florea_2005} and PCR\footnote{For historical reasons, PCR presented here was called PCR5 in our previous papers \cite{Smarandache_2005,Smarandache_2005c} since it results from a step-by-step improvement of a very simple PCR rule proposed in \cite{Smarandache_2005d}.} (Proportional Conflict Redistribution rule) \cite{Smarandache_2005,Smarandache_2005c} which are new efficient alternatives to Dempster's rule. The ACR is a mixing of the conjunctive and the disjunctive rules based on the distribution of the conflict according to a new choice of weighting coefficients. Using the ACR, a partial positive reinforcement of the belief can be observed for the focal elements commun to all the bbas to combine. The PCR redistributes the partial conflicting masses to the elements involved in the partial conflicts only, considering the conjunctive normal form of the partial conflicts. We restrict here our presentation to the simple case of the combination of two independent sources of evidence working on Shafer's model for the frame $\Theta=\{\theta_1, \ldots,\theta_n\}$, $n>1$ (finite set of exhaustive and exclusive hypotheses), although the extensions to the free and hybrid model of the DSmT can easily be obtained.. We assume the reader is already familiar with DST, with classical belief functions and Dempster's rule. Foundations of DST and its recent advances can be found in \cite{Shafer_1976,Yager_1994} while foundations of DSmT and its first applications can be found in \cite{DSmTBook_2004a}. In the next section we briefly remind only basics on DST and DSmT and the major fusion rules to make this paper self-consistent for the evaluation of simulation results. Section \ref{sec:ACR} is devoted to ACR while Section \ref{sec:PCR} is devoted to PCR. Section \ref{sec:examples} presents and compares simulation results and then we conclude in Section \ref{sec:conclusion}.

\section{Basics of DST, DSmT and fusion}
\label{sec:basics}

\subsection{Power set and hyper-power set}

Let $\Theta=\{\theta_{1},\ldots,\theta_{n}\}$ be a finite set (called frame) of $n$ exhaustive elements .
The free Dedekind's lattice denoted {\it{hyper-power set}}  $D^\Theta$ \cite{DSmTBook_2004a} is defined as 
\begin{enumerate}
\item $\varnothing, \theta_1,\ldots, \theta_n \in D^\Theta$.
\item  If $A, B \in D^\Theta$, then $A\cap B$ and $A\cup B$ belong to $D^\Theta$.
\item No other elements belong to $D^\Theta$, except those obtained by using rules 1 or 2.
\end{enumerate}
If $\vert\Theta\vert=n$, then $\vert D^\Theta\vert \leq 2^{2^n}$. Since for any finite set $\Theta$, $\vert D^\Theta\vert  \geq \vert 2^\Theta\vert $, we call $D^\Theta$ the  {\it{hyper-power set}} of $\Theta$.  The {\it{free DSm model}} $\mathcal{M}^f(\Theta)$ is based on $D^\Theta$ and allows to work with vague concepts which exhibit a continuous and relative intrinsic nature. Such concepts cannot be precisely refined in an absolute interpretation because of the unreachable universal truth. Shafer's model, denoted $\mathcal{M}^0(\Theta)$, assumes that all elements $\theta_{i}\in\Theta$, $i=1,\ldots,n$ are truly exclusive. In this case, all intersections involved in elements of $D^\Theta$ become empty and $D^\Theta$ reduces to classical power set denoted $2^\Theta$ \cite{Shafer_1976}. Between the free-DSm model and the Shafer's model, there exists a wide class of fusion problems represented in term of the DSm hybrid models where $\Theta$ involves both fuzzy continuous hypothesis and discrete hypothesis. Each hybrid fusion problem is then characterized by a proper hybrid DSm model $\mathcal{M}(\Theta)$ with $\mathcal{M}(\Theta)\neq\mathcal{M}^f(\Theta)$ and $\mathcal{M}(\Theta)\neq \mathcal{M}^0(\Theta)$.
The main differences between DST and DSmT are (1) the model on which one works with, and (2) the choice of the combination rule. We use here the generic notation $G$ for denoting either $D^\Theta$ (when working in DSmT) or $2^\Theta$ (when  working in DST). We denote $G^\ast$ the set $G$ from which the empty set is excluded $(G^\ast = G \setminus \{\varnothing)\})$.

\subsection{Basic belief functions}

A basic belief assignment (bba), called also belief mass, is defined as a mapping function $m(.): G \rightarrow [0,1]$ provided by a given source of evidence $\mathcal{B}$ satisfying 
\begin{equation}
m(\varnothing)=0 \qquad \text{and}\qquad \sum_{A\in G} m(A) = 1 
\end{equation}
The elements of $G$ having a strictly positive mass are called focal elements of $\mathcal{B}$. Let $\mathcal{F}$ be the set of focal elements of $m(.)$. In the DST framework $G$ can only be $2^\Theta$, while in the DSmT framework $G$ can be $D^\Theta$, a restricted-$D^\Theta$ given by some integrity constraints, or $2^\Theta$ and thus, we talk about the free model, the hybrid model or Shafer's model.

\subsection{Brief review of main fusion rules}

A wide variety of combination rules exists and a review and classification is proposed for example in \cite{Sentz_2002}, where the rules are analyzed according to their algebraic properties as well as on different examples. A recent review of main fusion rules can also be found in \cite{Smarandache_2005b,Smets_2005}. To simplify the notations, we consider only two independent sources of evidence $\mathcal{B}_1$ and $\mathcal{B}_2$ over the same frame $\Theta$ with their corresponding bbas $m_1(.)$ and $m_2(.)$. Even if the general case of $N$ different sources is defined it is not considered in this paper. Most of the fusion operators proposed in the literature so far use either the conjunctive operator, the disjunctive operator or a particular combination of them. These operators are respectively defined $\forall A\in G$, by
\begin{equation}
\label{eq:comb_conj}
m_{\wedge}(A)=(m_1 \wedge m_2) (A) =  \sum_{\substack{X,Y\in G\\ X \cap Y = A}} m_1(X)m_2(Y)
\end{equation}
\begin{equation}
\label{eq:comb_disj} 
m_{\vee}(A)=(m_1 \vee m_2) (A) =  \sum_{\substack{X,Y\in G\\ X \cup Y = A}} m_1(X)m_2(Y)
\end{equation}
\noindent
The {\it{degree of conflict}} between the sources $\mathcal{B}_1$ and $\mathcal{B}_2$ is defined by 
 \begin{equation}
 k_{12}\triangleq m_{\wedge}^{12}(\varnothing)=\displaystyle
\sum_{\substack{X,Y\in G\\ X\cap Y=\varnothing}} m_{1}(X) m_{2}(Y)
 \end{equation}
 \noindent
If $k_{12}$ is close to $0$, the bbas $m_1(.)$ and $m_2(.)$ are almost not in conflict, while if $k_{12}$ is close to $1$, the bbas are almost in total conflict. Next, we briefly review the main common fusion rules encountered in the literature and used in engineering applications.\\

\noindent
$\bullet$ {\bf{Dempster's rule} \cite{Dempster_1968}} : This combination rule has been proposed by Dempster. We assume (without loss of generality) that the sources of evidence are equally reliable. Otherwise a discounting preprocessing is first applied. It is defined on $G=2^\Theta$ by forcing $m_{DS}(\varnothing)\triangleq 0$ and $\forall A \in G^\ast$ by
\begin{equation}
m_{DS}(A) = \frac{1}{1-k_{12}} m_{\wedge}(A)=\frac{ m_{\wedge}(A)}{1-m_{\wedge}(\varnothing)}
\label{eq:DSR}
 \end{equation}
When $k_{12}=1$, this rule cannot be used. Dempster's rule of combination can be directly extended for the combination of $N$ independent and equally reliable sources of evidence and its major interest comes essentially from its commutativity and associativity properties \cite{Shafer_1976}. Dempster's rule corresponds to the normalized conjunctive rule by uniformly reassigning the mass of total conflict onto all focal elements through the conjunctive operator. The non normalized version of the Dempster's rule corresponds to the Smet's fusion rule in the TBM framework working under an open-world assumption, \ie $m_S(\varnothing)=k_{12}$ and $\forall A \in G^\ast$, $m_S(A)=m_{\wedge}(A)$.\\

\noindent
$\bullet$ {\bf{Yager's rule}} \cite{Yager_1983, Yager_1985,Yager_1987}: Yager admits that in case of conflict Dempster's rule provides counter-intuitive results. Thus, $k_{12}$ plays the role of an absolute discounting term added to the weight of ignorance. The commutative and quasi-associative\footnote{quasi-associativity was defined by Yager in \cite{Yager_1987}} Yager's rule is given by $m_Y(\varnothing)=0$ and $\forall A \in G^\ast$ by
\begin{equation}
\begin{cases}
m_Y(A)=m_{\wedge}(A) \\
m_Y(\Theta)=m_{\wedge}(\Theta) + m_{\wedge}(\varnothing)
\end{cases}
\label{eq:YagerRule}
\end{equation}

\noindent
$\bullet$ {\bf{Dubois \& Prade's rule}} \cite{Dubois_1988}: This rule supposes that the two sources are reliable when they are not in conflict and at least one of them is right when a conflict occurs. Then if one believes that a value is in a set $X$ while the other believes that this value is in a set $Y$, the truth lies in $X\cap Y$ as long $X\cap Y\neq \varnothing$. If $X\cap Y=\varnothing$, then the truth lies in $X\cup Y$. According to this principle, the commutative and quasi-associative Dubois \& Prade hybrid rule of combination, which is a reasonable trade-off between precision and reliability, is defined by $m_{DP}(\varnothing)=0$ and $\forall A \in G^\ast$ by
\begin{equation}
m_{DP}(A)=m_{\wedge}(A)  + \sum_{\substack{X,Y\in G\\ X\cup Y=A \\X\cap Y=\varnothing}} m_1(X)m_2(Y)
\label{eq:DuboisRule}
\end{equation}

\noindent
$\bullet$ {\bf{Inagaki's rule}} \cite{Inagaki_1991}: Inagaki proposed a very general formalism for all fusion rules which distributes the mass of the empty set after the conjunctive combination of $m_1(.)$ and $m_2(.)$. Inagaki's rule is given by $m_{Ina}(\varnothing)=0$ and $ \forall A \in G^\ast$ by
\begin{equation}
m_{Ina}(A) =m_{\wedge}(A) + w_m(A)m_{\wedge}(\varnothing)
\label{eq:CEV}
\end{equation}
\noindent with $w_m(A)\in[0,1]$, $\forall A \in G^*$ such that $\sum_{A\in G} w_m(A)=1$. It can be shown in \cite{Lefevre_2002,DSmTBook_2004a} that all previous combination rules (Dempster, Yager, Dubois \& Prade, Smets) can be obtained from Inagaki's formula \eqref{eq:CEV} with a proper choice of weighting factors $w_m(.)$. Inagaki also derived from \eqref{eq:CEV} a particular class of combination rules for which the ratio between the mass of any two subsets $A$ and $B$ (different from the frame $\Theta$) must be the same before and after the distribution of the mass of the empty set (see \cite{Inagaki_1991} for more details).\\

\noindent
$\bullet$ {\bf{Classic DSm fusion rule (DSmC)}} \cite{DSmTBook_2004a}:
Within the DSmT framework and when the free DSm model $\mathcal{M}^f(\Theta)$ holds, the conjunctive consensus, called the DSm classic rule (we will use the acronym DSmC in the sequel), is performed on $G = D^\Theta$. DSmC of two independent\footnote{While independence is a difficult concept to define in all theories managing epistemic uncertainty, we consider that two sources of evidence are independent (\ie distinct and noninteracting) if each leaves one totally ignorant about the particular value the other will take.} sources associated with $m_{1}(.)$ and $m_{2}(.)$ is thus given by \eqref{eq:comb_conj}. Since $G$ is closed under $\cup$ and $\cap$ set operators, DSmC guarantees that $m(.)$ is a proper belief assignment, \ie $m(.): G \rightarrow [0,1]$. DSmC is commutative, associative and can always be used for the fusion of sources involving fuzzy concepts whenever the free DSm model holds. This rule is directly and easily extended for the combination of $s > 2$ independent sources \cite{DSmTBook_2004a}.\\

\noindent
$\bullet$ {\bf{Hybrid DSm fusion rule (DSmH)}} \cite{DSmTBook_2004a}: DSmH generalizes DSmC and is no longer equivalent to Dempster's rule. DSmH is actually a direct extension of Dubois \& Prade's rule \cite{Dubois_1988} from the power-set $2^\Theta$ to the constrained hyper-power set $D^\Theta$ to take into account the possible dynamicity of the frame $\Theta$. It works for any models (the free DSm model, Shafer's model or any other hybrid models) when manipulating {\it{precise}} generalized (\ie defined over $D^\Theta$) or eventually classical (\ie defined over $2^\Theta$) basic belief assignments. A complete description of this combination rule is given in \cite{DSmTBook_2004a}.

\section{The Adaptive Combination Rule}
\label{sec:ACR}

A new class of combination rules - a mixing between the conjunctive rule $\wedge$ and the disjunctive rule $\vee$ (defined respectively by \eqref{eq:comb_conj} and \eqref{eq:comb_disj}) was proposed for evidence theory in \cite{Florea_2004}. Hence, we assume Shafer's model and thus work on the power set $(G = 2^\Theta)$. The {\it{generic}} Adaptive Combination Rule (ACR) between $m_1(.)$ and $m_2(.)$ is defined by $m_{ACR}(\varnothing) =0$ and $\forall A \in G^\ast$ by
\begin{equation}
m_{ACR} (A) = \alpha(k_{12}) m_{\vee}(A) + \beta(k_{12}) m_{\wedge}(A)
\label{eq:comb_adaptive}
\end{equation}
\noindent where $\alpha$ and $\beta$ are functions of the conflict $k_{12}=m_{\wedge}(\varnothing)$ from $[0, 1]$ to $[0, + \infty[$. $m_{ACR}(.)$ must be a normalized bba (we assume here a closed world) and a desirable behaviour of ACR is to act more like the disjunctive rule $\vee$ whenever $k_{12}$ is close to $1$ (\ie at least one source is unreliable), while it should act more like the conjunctive rule $\wedge$, when $k_{12}$ becomes close to $0$ (\ie both sources are reliable). Hence, the three following conditions should be satisfied by the weighting functions $\alpha$ and $\beta$ : 
\begin{enumerate}
\item[(C1)] $\alpha$ is increasing with $\alpha (0) = 0$ and $\alpha (1) = 1$; 
\item[(C2)] $\beta$ is decreasing with $\beta (0) = 1$ and $\beta (1) = 0$.
\item[(C3)] $\alpha (k_{12}) = 1 - (1 - k_{12}) \beta (k_{12})$
\end{enumerate}
\noindent
The Condition (C3) is given by the necessity of the $m_{ACR}$ to be a bba ($\sum_{A \in G} m_{ACR}(A)=1$).

\noindent
It has been shown however in \cite{Florea_2004} that (C1) is actually a direct consequence of (C2) and (C3) and becomes irrelevant. This class of ACR  can be stated from \eqref{eq:comb_adaptive}, for any function $\beta$ satisfying (C2) and for $\alpha$ given by condition (C3). \\

\noindent
Here are some important remarks on the class of ACR as presented in \cite{Florea_2004} :
\begin{enumerate}
\item The class of ACR is a particular case of Inagaki's general class of combination rules with weighting factors expressed as (see \cite{Florea_2004} for proof)
\begin{equation}
\begin{split}
\label{eq:adapt_weighting}
w_m(A) = \frac{1 - \beta(k_{12})}{k_{12}} [m_{\vee}(A) - m_{\wedge}(A)] \\
+ \beta(k_{12}) m_{\vee}(A)
\end{split}
\end{equation}
\item $w_m(A)$ drawn from ACR can be negative in \eqref{eq:adapt_weighting}, \ie $w_m(A) < 0$ $\forall A\in G^\ast$ such that
\begin{equation*}
m_{\wedge}(A) > m_{\vee}(A)[ 1 + \frac{k_{12} \beta(k_{12})}{1 - \beta(k_{12})}]
\end{equation*}
Thus, ACR defined previously may be viewed as an extension of Inagaki's rules \eqref{eq:CEV}.
\item The ACR creates a bba with focal elements chosen from the focal elements produced by the conjunctive or the disjunctive combination rules ($\mathcal{F}_{ACR} = \mathcal{F}_{1 \wedge 2} \cup \mathcal{F}_{1 \vee 2}$). Inagaki's general class of combination rules can distribute the mass of the empty set ($k_{12}$) to any subset of $\Theta$, thus is more general than the ACR, but this is not necessarily an asset.
\item The combination of $m_1(.)$ and $m_2(.)$ using the ACR leads to a {\em partial positive reinforcement of the belief} for the focal elements common to both $\mathcal{F}_1$ and $\mathcal{F}_2$.\\
\end{enumerate}

\noindent
It can easily be shown that the ACR preserves the neutral impact of the vacuous belief in the fusion processes. \\

\noindent A symmetric ACR (SACR for short), \ie with symmetric weightings for $m_{\wedge}(.)$ and $m_{\vee}(.)$, such that $\alpha(k_{12}) = 1 - \beta(1 - k_{12})$, was also introduced in \cite{Florea_2004}. This choice was imposed by a particular behaviour for the ACR. The SACR is defined by $m_{SACR}(\varnothing)=0$ and $\forall A \in G^\ast$ by
\begin{equation}
\label{eq:comb_adapt_5}
m_{SACR}(A)= \alpha_0(k_{12}) m_{\vee}(A) + \beta_0(k_{12}) m_{\wedge}(A)
\end{equation}
\noindent
where
\begin{equation}
\begin{split}
\alpha_0(k_{12}) = \frac{\D k_{12}}{\D 1 - k_{12} + k_{12}^2 }\\
\beta_0(k_{12}) = \frac{\D 1 - k_{12}}{\D 1 - k_{12} + k_{12}^2 }
\label{eq:alpha0_beta0} 
\end{split}
\end{equation}
\noindent 
In \cite{Florea_2004}, the authors show the uniqueness of SACR.

\section{Proportional Conflict Redistribution}
\label{sec:PCR}

\subsection{Principle of PCR}

Instead of applying a direct transfer of partial conflicts onto partial uncertainties as with DSmH rule, the idea behind the Proportional Conflict Redistribution (PCR) rule \cite{Smarandache_2005,Smarandache_2005c} is to transfer conflicting masses (total or partial) proportionally to non-empty sets involved in the model according to all integrity constraints. 
\noindent
The general principle of PCR rules is to :
\begin{enumerate}
\item calculate the conjunctive rule of the belief masses of sources ;
\item calculate the total or partial conflicting masses ;
\item redistribute the conflicting mass (total or partial) proportionally on non-empty sets involved in the model according to all integrity constraints. 
\end{enumerate}
 
The way the conflicting mass is redistributed yields actually to five versions of PCR rules, denoted PCR1-PCR5 which have been presented in \cite{Smarandache_2005,Smarandache_2005c}. These PCR fusion rules work for any degree of conflict $k_{12} \in [0, 1]$ or $k_{12\ldots s} \in [0, 1]$, for any DSm models (Shafer's model, free DSm model or any hybrid DSm model) and both in DST and DSmT frameworks for static or dynamical fusion problematics. We present below only the most sophisticated proportional conflict redistribution rule (corresponding to PCR5 in \cite{Smarandache_2005,Smarandache_2005c} but denoted here just PCR) since this rule is what we feel the most efficient PCR fusion rule developed so far.

\subsection{Explicit formula for PCR for two sources}

The PCR rule redistributes the partial conflicting mass to the elements involved in the partial conflict, considering the conjunctive normal form of the partial conflict. PCR is what we think the most mathematically exact redistribution of the conflicting mass obatined after the conjunctive rule. PCR rule preserves the neutral impact of the vacuous belief assignment because the mass of the focal element $\Theta$ cannot be involved in the conflict. Since $\Theta$ is a neutral element for the intersection (conflict), $\Theta$ gets no mass after the redistribution of the conflicting mass. We have also proven the continuity property of the PCR result with continuous variations of bbas to combine in \cite{Smarandache_2005}. PCR rule for two sources is given by:  $m_{PCR}(\varnothing)=0$ and $\forall X\in G^\ast$

\begin{multline}
m_{PCR}(X)=m_{\wedge}(X) +\sum_{\substack{Y\in G\setminus\{X\} \\ c(X\cap Y)=\varnothing}} 
[\frac{m_1(X)^2m_2(Y)}{m_1(X)+m_2(Y)} +\\
 \frac{m_2(X)^2 m_1(Y)}{m_2(X)+m_1(Y)}]
   \label{eq:PCR5}
 \end{multline}
\noindent
where $c(X)$ is the canonical form\footnote{The canonical form is the conjunctive normal form, also known as conjunction of disjunctions in Boolean algebra, which is unique and is its simplest form. For example if $X=(A\cap B)\cap (A\cup B\cup C)$, 
$c(X)=A\cap B$.}  (conjunctive normal) of $X$ and where all denominators are {\it{different from zero}}. If a denominator is zero, that fraction is discarded. The general PCR formula for $s\geq 2$ sources is given in \cite{Smarandache_2005}.


\section{Illustrative examples}
\label{sec:examples}

\subsection{A simple two-source example}

\begin{itemize}
\item {\bf{Example 1}}: Let us take $\Theta=\{A, B\}$ of exclusive elements (Shafer's model), and the following bbas:
\begin{center}
\begin{tabular}[h]{|c|ccc|}
\hline
 & $A$ & $B$ & $ A\cup B$\\
 \hline
 $m_1(.)$ & 0.6 & 0 & 0.4 \\
 \hline
 $m_2(.)$ & 0 & 0.3 & 0.7 \\
 \hline
 \hline
 $m_{\wedge}(.)$ & 0.42 & 0.12 & 0.28 \\
 \hline
\end{tabular}
\end{center}
The conflicting mass is $k_{12}=m_{\wedge}(A\cap B)$ and equals $m_1(A)m_2(B)+m_1(B)m_2(A)=0.18$.
Therefore $A$ and $B$ are the only focal elements involved in the conflict. Hence according to the PCR hypothesis only $A$ and $B$ deserve a part of the conflicting mass and $A\cup B$ does not deserve. With PCR, one redistributes the conflicting mass $k_{12}=0.18$ to $A$ and $B$ proportionally with the masses $m_1(A)$
and $m_2(B)$ assigned to $A$ and $B$ respectively. Let $x$ be the conflicting mass to be redistributed
to $A$, and $y$ the conflicting mass redistributed to $B$, then
$$\frac{x}{0.6}=\frac{y}{0.3}=\frac{x+y}{0.6+0.3}=\frac{0.18}{0.9}=0.2$$
\noindent
hence $x = 0.6\cdot 0.2 = 0.12$, $y = 0.3\cdot 0.2 = 0.06$. Thus, the final result using the PCR rule is
%
\begin{equation*}
\begin{cases}
 m_{PCR}(A)=0.42 + 0.12 = 0.54 \\
 m_{PCR}(B)= 0.12 + 0.06 = 0.18 \\
 m_{PCR}(A\cup B)= 0.28
\end{cases}
\end{equation*}

\noindent With SACR, $\alpha_0(0.18)\approx 0.211$ and $\beta_0(0.18)\approx 0.962$ and therefore
\begin{equation*}
\begin{cases}
 m_{SACR}(A)=\alpha_0\cdot 0 + \beta_0\cdot 0.42 \approx  0.404 \\
 m_{SACR}(B)=\alpha_0\cdot 0 + \beta_0\cdot 0.12 \approx 0.116\\
 m_{SACR}(A\cup B)= \alpha_0\cdot 1 + \beta_0\cdot 0.28\approx 0.480
\end{cases}
\end{equation*}

We summarize in the following table the previous results and the results obtained from other rules presented in Section \ref{sec:basics} (three decimals approximations).
\begin{center}
\begin{tabular}[h]{|l||ccc|}
\hline
  & $A$ & $B$ & $A\cup B$\\
 \hline
  $m_{DS}$ & 0.512 & 0.146 & 0.342 \\
 $m_{DP}$ & 0.420 & 0.120 & 0.460 \\
 $m_{DSmH}$ & 0.420 & 0.120 & 0.460 \\
 $m_{Y}$ & 0.420 & 0.120 & 0.460 \\
 $m_{Ina}$ & 0.560 & 0.160 & 0.280 \\
 $m_{SACR}$ & 0.404 & 0.116 & 0.480 \\
 $m_{PCR}$ & 0.540 & 0.180 &  0.280 \\
\hline
\end{tabular}
\end{center}
\noindent

Note that in this particular 2D case DSmH, Dubois \& Prade's and Yager's rules coincide. They do not coincide in general when $|\Theta | > 2$. ACR provides very close results as DSmH, DP and Y. Inagaki's optimal combination rule was used in this example (see \cite{Inagaki_1991} for more details). Smets' and DSmC rules have not been included in this table since they are based on different models (open-world and free-DSm model respectively). They cannot be compared formally to the other rules since Shafer's model does not hold anymore. Within DSmC one keeps separately all the masses committed to partial conflicts while within Smets' rule all partial conflicts are reassigned to the empty set interpreted as all missing hypotheses.

\item {\bf{Example 2}}: Let us modify a little bit the previous example and consider now the following bbas:
\begin{center}
\begin{tabular}[h]{|c|ccc|}
\hline
 & $A$ & $B$ & $ A\cup B$\\
 \hline
 $m_1(.)$ & 0.6 & 0 & 0.4 \\
 \hline
 $m_2(.)$ & 0.2 & 0.3 & 0.5 \\
 \hline
 \hline
 $m_{\wedge}(.)$ & 0.50 & 0.12 & 0.20 \\
 \hline
\end{tabular}
\end{center}
\noindent The conflicting mass $k_{12}=m_{\wedge}(A\cap B)$ as well as the distribution coefficients $x$ and $y$ for the PCR rule and the weighting coefficients $\alpha_0\approx 0.211$ and $\beta_0\approx 0.962$ for the SACR rule remain the same as in the previous example. Thus, the result obtained using the PCR rule is:
\begin{equation*}
\begin{cases}
 m_{PCR}(A)=0.50 + 0.12 = 0.620\\
 m_{PCR}(B)= 0.12 + 0.06 = 0.180 \\
 m_{PCR}(A\cup B)=  0.20 +0 = 0.200\\
\end{cases}
\end{equation*}
Using SACR rule, the result of the combination is:
\begin{equation*}
\begin{cases}
 m_{SACR}(A)=\alpha_0\cdot 0.12 + \beta_0\cdot 0.50 \approx  0.506\\
 m_{SACR}(B)=\alpha_0\cdot 0 + \beta_0\cdot 0.12 \approx 0.116 \\
 m_{SACR}(A\cup B)= \alpha_0\cdot 0.88 + \beta_0\cdot 0.20\approx 0.378
\end{cases}
\end{equation*}

All fusion rules based on Shafer's model are used in this example and the results are presented in the following table (three decimals approximations).
\begin{center}
\begin{tabular}[h]{|l||ccc|}
\hline
  & $A$ & $B$ & $A\cup B$\\
 \hline
  $m_{DS}$ & 0.609 & 0.146 & 0.231 \\
 $m_{DP}$ & 0.500 & 0.120 & 0.380 \\
 $m_{DSmH}$ & 0.500 & 0.120 & 0.380 \\
 $m_{Y}$ & 0.500 & 0.120 & 0.380 \\
 $m_{Ina}$ & 0.645 & 0.155 & 0.200 \\
 $m_{SACR}$ & 0.506 & 0.116 & 0.378 \\
 $m_{PCR}$ & 0.620 & 0.180 &  0.200 \\
\hline
\end{tabular}
\end{center}
\noindent
In this example SACR is very close to DP, Y and DSmH rules while PCR is more close to DS and Inagaki's rule.

\item {\bf{Example 3}}:
Let's go further modifying this time the previous example and considering the following bbas:
\begin{center}
\begin{tabular}[h]{|c|ccc|}
\hline
 & $A$ & $B$ & $ A\cup B$\\
 \hline
 $m_1(.)$ & 0.6 & 0.3 & 0.1 \\
 \hline
 $m_2(.)$ & 0.2 & 0.3 & 0.5 \\
 \hline
 \hline
 $m_{\wedge}(.)$ & 0.44 & 0.27 & 0.05 \\
 \hline
\end{tabular}
\end{center}
The conflicting mass $k_{12}=m_{\wedge}(A\cap B)=0.24 = 0.18+0.06 = m_1(A)m_2(B)+m_1(B)m_2(A)$ is now different from the two previous examples, which means that $m_2(A) = 0.2$ and $m_1(B) =0.3$ did make an impact on the conflict. Therefore $A$ and $B$ are the only focal elements involved in the conflict and thus only $A$ and $B$ deserve a part of the conflicting mass. PCR redistributes the partial conflicting mass 0.18 to $A$ and $B$ proportionally with the masses $m_1(A)$ and $m_2(B)$ (let $x_1$ and $y_1$ be the conflicting mass to be redistributed to $A$ and $B$, respectively) and also the partial conflicting mass 0.06 to $A$ and $B$ proportionally with the masses $m_2(A)$ and $m_1(B)$ (let $x_2$ and $y_2$ be the conflicting mass to be redistributed to $A$ and $B$, respectively). The distribution coefficients $x_1$ and $y_1$ are those computed in the two previous examples ($x_1 = 0.12$ and $y_1 = 0.06$). To compute the second pair of distribution coefficients, one has: 
$$\frac{x_2}{0.2}=\frac{y_2}{0.3}=\frac{x_2+y_2}{0.2+0.3}=\frac{0.06}{0.5}=0.12$$ whence $x_2 = 0.2\cdot 0.12 = 0.024$ and $y_2 = 0.3\cdot 0.12 = 0.036$. 
Thus, the result obtained using the PCR rule is:
\begin{equation*}
\begin{cases}
 m_{PCR}(A)=0.44 + 0.12 + 0.024 = 0.584 \\
 m_{PCR}(B)= 0.27 + 0.06 + 0.036 = 0.366 \\
 m_{PCR}(A\cup B)= 0.05 + 0 = 0.05
\end{cases}
\end{equation*}
\noindent

Since the conflict is $k_{12}=0.24$, the weighting coefficients for the SACR become $\alpha_0\approx 0.294$ and $\beta_0\approx 0.930$ and the result is:
\begin{equation*}
\begin{cases}
 m_{SACR}(A)=\alpha_0\cdot 0.12 + \beta_0\cdot 0.44 \approx  0.445\\
 m_{SACR}(B)=\alpha_0\cdot 0.09 + \beta_0\cdot 0.27 \approx 0.277\\
 m_{SACR}(A\cup B)= \alpha_0\cdot 0.79 + \beta_0\cdot 0.05\approx 0.278
\end{cases}
\end{equation*}

All fusion rules based on Shafer's model are used in this example and the results are presented in the following table (three decimals approximations).
\begin{center}
\begin{tabular}[h]{|l||ccc|}
\hline
  & $A$ & $B$ & $A\cup B$\\
 \hline
  $m_{DS}$ & 0.579 & 0.355 & 0.066 \\
 $m_{DP}$ & 0.440 & 0.270 & 0.290 \\
 $m_{DSmH}$ & 0.440 & 0.270 & 0.290 \\
 $m_{Y}$ & 0.440 & 0.270 & 0.290 \\
 $m_{Ina}$ & 0.588 & 0.362 & 0.050 \\
 $m_{SACR}$ & 0.445 & 0.277 & 0.278 \\
 $m_{PCR}$ & 0.584 & 0.366 &  0.050 \\
\hline
\end{tabular}
\end{center}
\noindent

One clearly sees that $m_{DS}(A\cup B)$ gets some mass from the conflicting mass although $A\cup B$ does not deserve any part of the conflicting mass (according to PCR hypothesis) since $A\cup B$ is not involved in the conflict (only $A$ and $B$ are involved in the conflicting mass). Dempster's rule appears to us less exact than PCR and Inagaki's rules. PCR result is very close to Inagaki's result but upon our opinion is more exact. SACR follows behaviours of DP, Y and DSmH.
\end{itemize}

\begin{figure*}[!ht]
\begin{center}
\includegraphics[height=5cm]{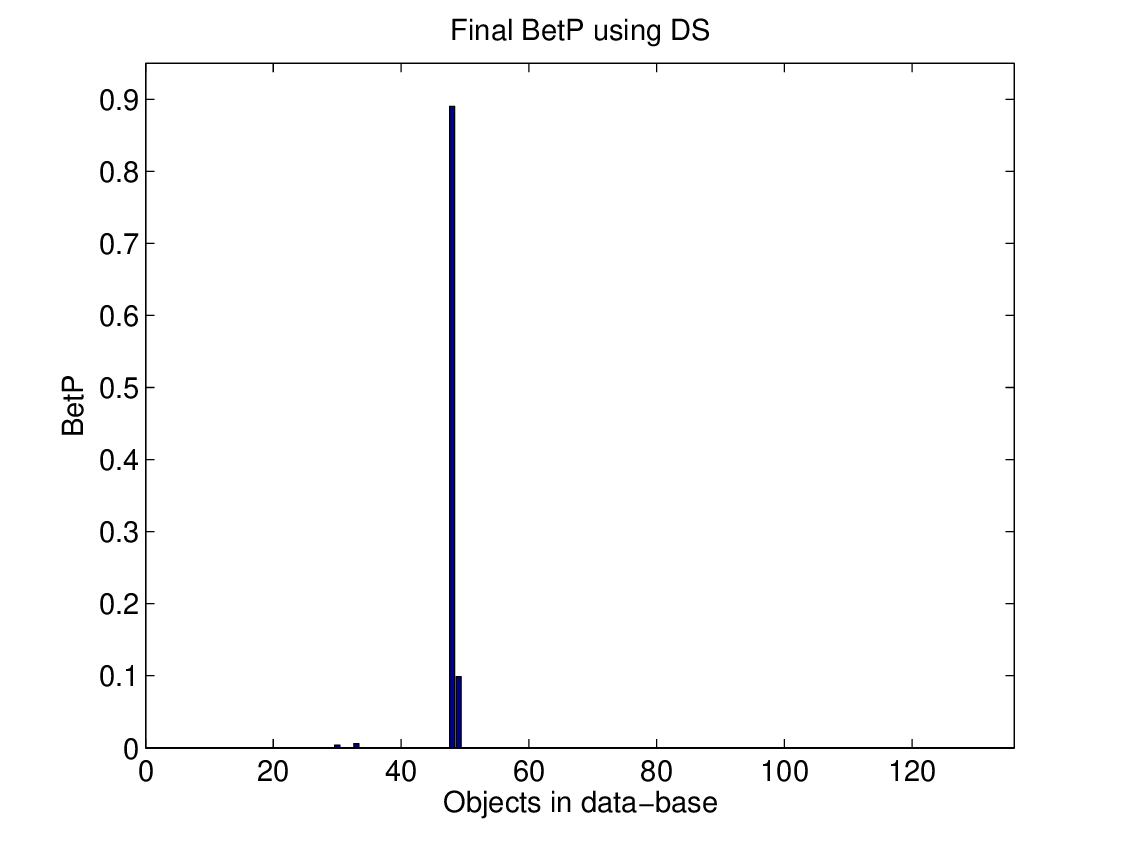} \hskip 1cm
\includegraphics[height=5cm]{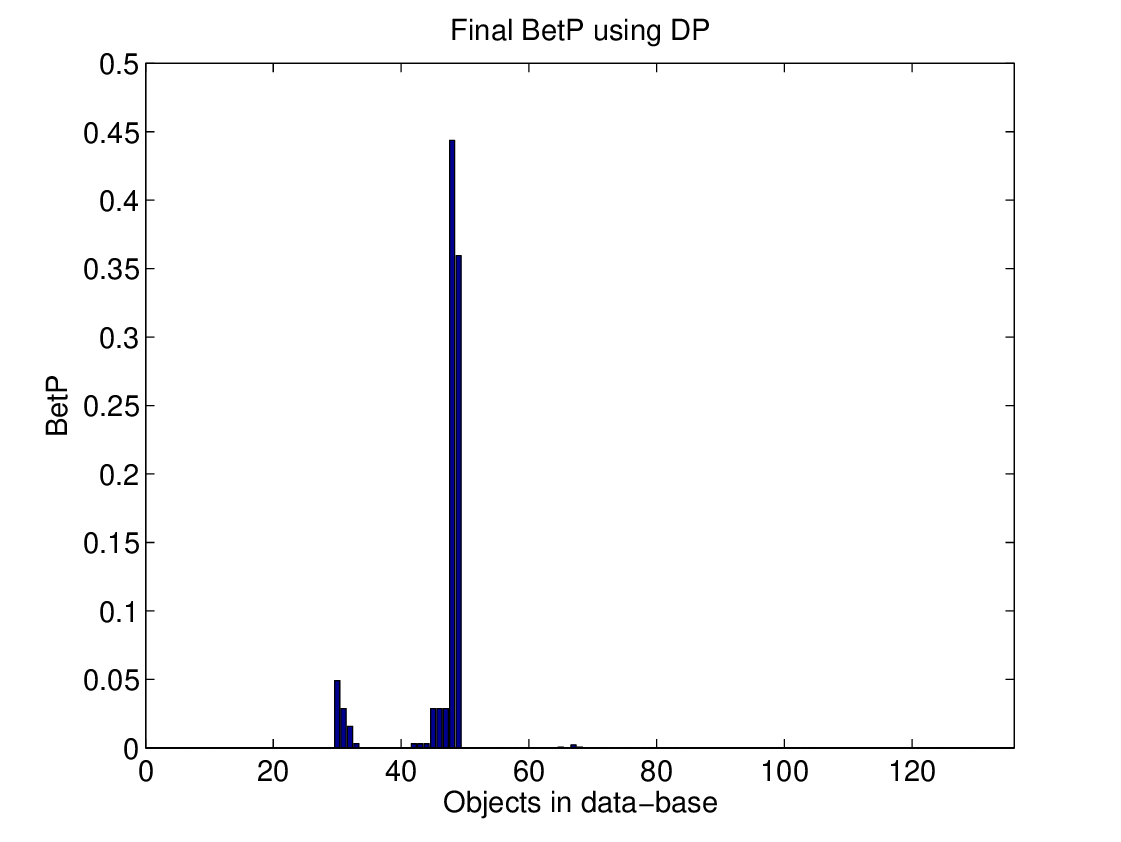} \\
\includegraphics[height=5cm]{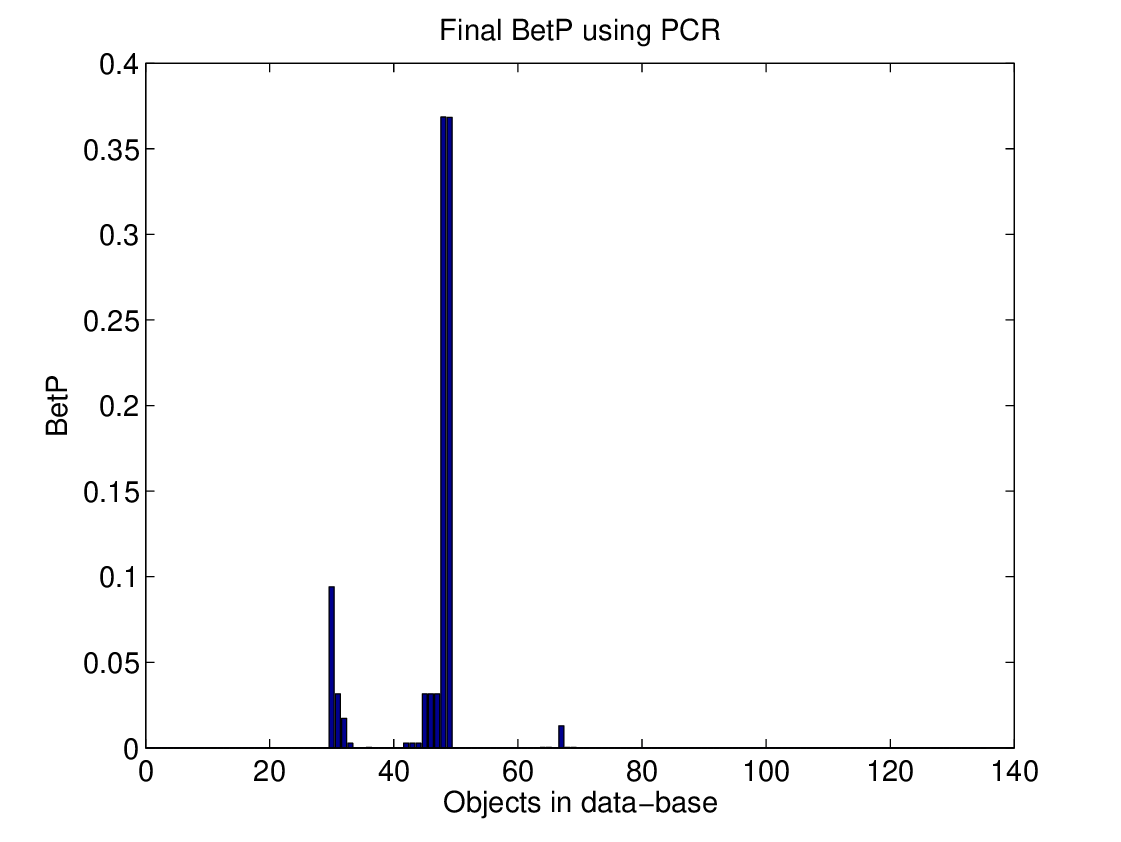}\hskip 1cm
\includegraphics[height=5cm]{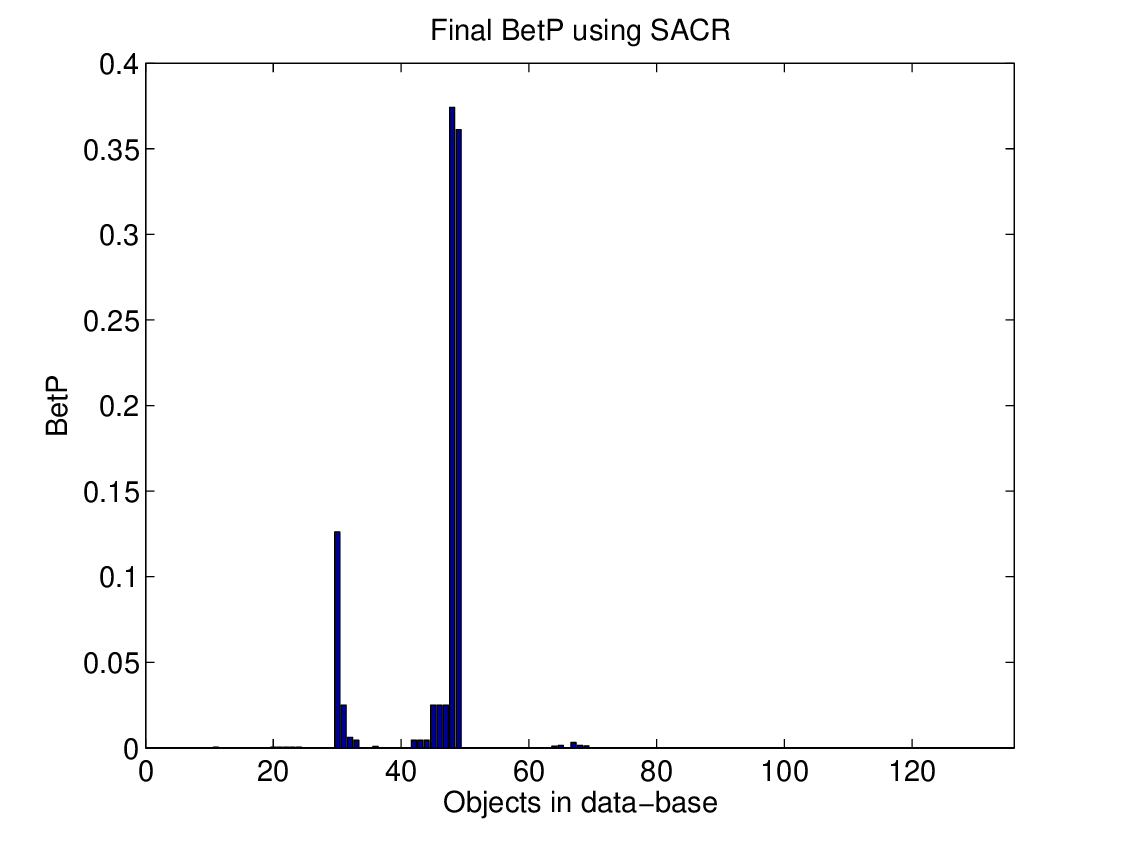}
\end{center}
\vskip -0.7cm
\caption{Test scenario of target identification fusion ESM reports with a probability of false alarm of 0.3.}
\vskip -0.2cm
\label{fig:Final}
\end{figure*}

\subsection{Zadeh's example}

We compare here the solutions for the well-known Zadeh's example \cite{Zadeh_1979} provided by several fusion rules. A detailed presentation with more comparisons can be found in \cite{DSmTBook_2004a,Smarandache_2005}. Let us consider the frame $\Theta=\{A,B,C\}$, the Shafer's model and the two following belief assignments :
\begin{align*}
m_1(A)&=0.9 &\quad m_1(B)&=0 &\quad m_1(C)&=0.1\\
m_2(A)&=0 &\quad m_2(B)&=0.9 &\quad m_2(C)&=0.1
\end{align*}
The total conflicting mass is high since it is
$$m_1(A)m_2(B)+m_1(A)m_2(C)+m_2(B)m_1(C)=0.99$$

All fusion rules based on Shafer's model are used in this example and the results are presented in the following table
\begin{center}
\begin{tabular}[h]{|l||c|c|c|c|}
\hline
  & $m_{DS}$ & $m_{Y}$ & $m_{DP}$ & $m_{DSmH}$ \\
\hline
$C$ & 1 &0.01 & 0.01 & 0.01 \\
$A\cup B$ & 0 & 0 & 0.81 & 0.81\\
$A\cup C$ & 0 & 0 &  0.09 & 0.09\\
$B\cup C$ &  0 & 0 &  0.09 & 0.09\\
$A\cup B\cup C$ & 0 & 0.99 & 0 & 0 \\
\hline
\end{tabular}
\vskip 0.3cm
\begin{tabular}[h]{|l||c|c|c|}
\hline
  & $m_{Ina}$ & $m_{SACR}$ & $m_{PCR}$\\
\hline
$A$ & 0 & 0 & 0.486\\
$B$ & 0 & 0 & 0.486\\
$C$ & 1 & $\approx$ 0.0101 & 0.028\\
$A\cup B$ & 0 & $\approx$ 0.8099 & 0 \\
$A\cup C$ & 0 & $\approx$ 0.0900 & 0 \\
$B\cup C$ & 0 & $\approx$ 0.0900 & 0 \\
\hline
\end{tabular}
\end{center}

\noindent
We can see that Dempster's rule yields the counter-intuitive result (see justifications in \cite{Zadeh_1979,Dubois_1986c,Yager_1987,Voorbraak_1991,DSmTBook_2004a}) which reflects the minority opinion. The Dubois \& Prade's rule (DP) \cite{Dubois_1986c} based on Shafer's model provides in this Zadeh's example the same result as DSmH, because DP and DSmH coincide in all static fusion problems\footnote{DP rule has been developed for static fusion only while DSmH is more general since it works for any models as well as for static and dynamic fusion.}. SACR is very close to DP and DSmH since the conflict is close to 1 and the SACR acts more like the disjunctive rule (according to the definition hypothesis). PCR acts more like a mean operator over the two bbas and is similar to Murphy's rule \cite{Murphy_2000}. 

%

\subsection{Target classification example}

In this section we study a simple test scenario of target identification. Several pieces of evidential information coming from an ESM (Electronic Support Measures) analyzing a combat scene are sequentially combined at a fusion centre. The 135 targets to be potentially identified are listed in a Platform Data Base (PDB), according to 22 features. One of these features is the emitters on board for each target. Hence we have, $\Theta = \{\theta_1, \ldots, \theta_{135}\}$.

The following simulation test was randomly generated considering that the probability of false alarm of the ESM is $0.3$, which means that 3 times over 10 the emitter reported does not belong to the observed target. Object $\theta_{48}$ is the observed object and it is the ground truth. We built two sets of emitters - the set $X$ having only the emitters of $\theta_{48}$ and the set $Y$ having emitters used on objects similar to $\theta_{48}$ which are not in $X$. Randomly, we choose an emitter from $X$ 7 times of 10 and from $Y$ 3 times of 10. We generated 25 such emitters (the repetitions were allowed) and each piece of information was modelled in evidence theory using the following bba: $m_0(A)=0.8$ and $m_0(\Theta)=0.2$ where $A$ is the subset of $\Theta$ corresponding to the received information (\ie each element of $A$ owns the emitter reported by the ESM). The successive bbas $m_t$, $m_{t+1}$ are combined using different combination rules, and the decision on the most probable observed target is taken following the maximum of pignistic probability\footnote{The pignistic probability (BetP) was introduced by Smets in \cite{Smets_1990b}} criterion (max BetP) \cite{Smets_1990b}.

\begin{figure}[!ht]
\begin{center}
\includegraphics[height=6cm]{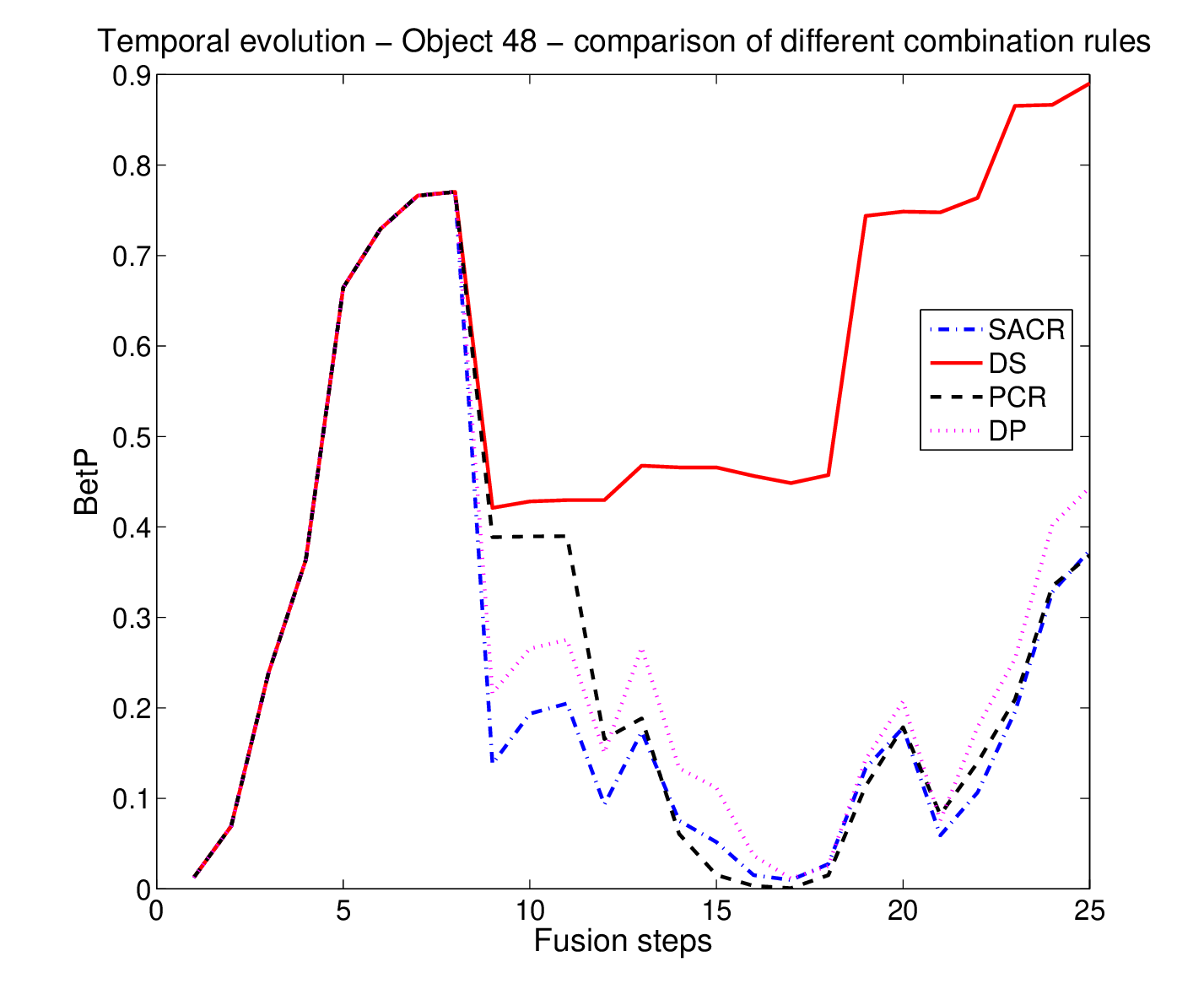} 
\end{center}
\vskip -0.7cm
\caption{Evolution of the pignistic probability of Singleton $\theta_{48}$ using different combination rules.}
\label{fig:Comparison_48}
\vskip -0.2cm
\end{figure}

The final pignistic probabilities obtained using DS, DP, PCR and SACR after 25 combination steps are shown in Figure \ref{fig:Final}. Figure \ref{fig:Comparison_48} shows a comparative temporal evolution of the pignistic probability of Singleton $\theta_{48}$ during the fusion process. In our data base, the difference between objects $\theta_{48}$ and $\theta_{49}$ is only given by one emitter. Thus, we expect Singleton $\theta_{48}$ to have a maximum BetP, but $\theta_{49}$ must be provided as a possible option for the identification. Using all four decision rules, we identify the Singleton $\theta_{48}$, which is our ground truth. However, using DS, the final BetP gives almost all credibility to Singleton $\theta_{48}$ without giving any chances to other Singletons, which is in opposition with our expectations. DP, SACR and PCR give to Singleton $\theta_{49}$ a high probability, which is however smaller than the probability of Singleton $\theta_{48}$. In the case of PCR, the choice between Singletons $\theta_{48}$ and $\theta_{49}$ is made with more difficulty than in the case of SACR and DP.

\section{Conclusion}
\label{sec:conclusion}

We discussed here two new and interesting combination rules for evidence theory: (1) the class of adaptive combination rules (ACR) with its particular case the symmetric adaptive combination rule (SACR) and (2) the proportional conflict redistribution rule (PCR).  PCR can also be used in DSm free model or in any hybrid model. These two new combination rules are able to cope with conflicting information contrary to the classic Dempster's rule. Some simple examples and a target classification example were presented to show their interest in defense application. Both SACR and PCR were compared to some classical rules showing their ability to combine high conflicting information in a new robust and better way than conventional rules used so far.


\end{document}